%% file: ijcai24.tex
\documentclass{article}
\usepackage{ijcai24}

\usepackage{times}
\usepackage{soul}
\PassOptionsToPackage{hyphens}{url}
\usepackage[hidelinks]{hyperref}
\usepackage[utf8]{inputenc}
\usepackage[small]{caption}
\usepackage{graphicx}
\usepackage{amsmath}
\usepackage{amsthm}
\usepackage{mathtools}
\usepackage{booktabs}
\usepackage{algorithm}
\usepackage[noend]{algpseudocode}
\usepackage[switch]{lineno}
\usepackage{amssymb}
\usepackage{xcolor}
\usepackage{cleveref}
\usepackage[most]{tcolorbox}
\usepackage{dsfont} 
\usepackage{multirow} 
\usepackage{multicol} 
\usepackage{subcaption}
\usepackage{wrapfig}
\usepackage{siunitx} 
\usepackage{pgfplots}
\pgfplotsset{compat=newest}

\input{defs}

\title{Human-Agent Cooperation in Games under Incomplete Information \\through Natural Language Communication\footnote{The source code and the full paper with appendix can be found at \url{https://github.com/vivianchen98/shared_control_language}.}}

\author{
Shenghui Chen$^1$\and
Daniel Fried$^2$\And
Ufuk Topcu$^1$\\
\affiliations
$^1$University of Texas at Austin,\\
$^2$Carnegie Mellon University\\
\emails
\{shenghui.chen, utopcu\}@utexas.edu,
dfried@cs.cmu.edu
}

\begin{document}
\maketitle

\begin{abstract}
Developing autonomous agents that can strategize and cooperate with humans under information asymmetry is challenging without effective communication in natural language.
We introduce a \textit{shared-control game}, where two players collectively control a \textit{token} in alternating turns to achieve a common objective under incomplete information.
We formulate a policy synthesis problem for an autonomous agent in this game with a human as the other player.
To solve this problem, we propose a communication-based approach comprising a language module and a planning module.
The language module translates natural language messages into and from a finite set of \textit{flags}, a compact representation defined to capture player intents.
The planning module leverages these flags to compute a policy using an \textit{asymmetric information-set Monte Carlo tree search with flag exchange} algorithm we present.
We evaluate the effectiveness of this approach in a testbed based on Gnomes at Night, a search-and-find maze board game.
Results of human subject experiments show that communication narrows the information gap between players and enhances human-agent cooperation efficiency with fewer turns.
\end{abstract}

\input{sections/1_introduction}
\input{sections/2_model_and_problem}
\input{sections/3_related}
\input{sections/4_method}
\input{sections/5_experiments}

\input{sections/6_conclusion}

\section*{Ethical Statement}
The user study was approved by the University of Texas at Austin Institutional Review Boards under ID STUDY00005476.

\section*{Acknowledgments}
This work was supported in part by the National Science Foundation (NSF) under grants 1652113, 271643-874F, and 2211432.
Additional support was provided by the Defense Advanced Research Projects Agency (DARPA) under Agreement No. HR00112490410.

\bibliographystyle{named}
\bibliography{ijcai24}

\onecolumn
\input{sections/supplementary}

\end{document}

%% file: defs.tex
\newcommand{\Scal}{\mathcal{S}}
\newcommand{\Acal}{\mathcal{A}}
\newcommand{\Tcal}{\mathcal{T}}
\newcommand{\Rcal}{\mathcal{R}}
\newcommand{\Fcal}{\mathcal{F}}

\newcommand{\human}{\textbf{H}}
\newcommand{\ego}{\textbf{E}}

\newcommand{\humanIt}{\textit{H}}
\newcommand{\agentComm}{\textit{A-Comm}}
\newcommand{\agentMute}{\textit{A-Mute}}

\definecolor{egoblue}{HTML}{4472C4} 
\definecolor{humangreen}{HTML}{70AD47}

\definecolor{humanhumangray}{HTML}{B7B7B7}
\definecolor{agentagentorange}{HTML}{F2A95D}
\definecolor{agentmutebrown}{HTML}{B88A6D}

\definecolor{pptblue}{HTML}{4472C4}

\theoremstyle{definition}

\newtheorem{definition}{Definition}
\newtheorem{problem}{Problem}

\newcommand{\argmax}[1]{\operatorname{arg\,max}\limits_{#1}}

\DeclareMathOperator{\subjectto}{s.t.}

\newcommand{\ucb}{\operatorname{ucb}}

  {\endquote}            

%% file: sections/1_introduction.tex
\section{Introduction}\label{sec:intro}

Developing autonomous agents capable of cooperative strategic planning in games under incomplete information is an important problem in human-agent interaction. Agents with such capabilities hold the potential to improve how they work with humans in a diverse range of settings, from virtual applications like strategy board games \cite{bard2020hanabi} and action video games \cite{carroll2019utility} to physical applications like assistive wheelchairs \cite{goil2013using}.

In games where the information available to each player is different or incomplete, cooperation between players is challenging without an effective exchange of information. This information asymmetry, often resulting from a partial observation of underlying game states or an incomplete grasp of overall game dynamics, can lead to actions that are either misaligned with the team's objectives or suboptimal for coordination. To overcome this challenge, it is crucial for players to engage in effective communication to exchange relevant information at runtime. If only autonomous agents are playing such games, they can accomplish this exchange in a structured manner, for example, with limited hints in the game of Hanabi \cite{bard2020hanabi}.

When an autonomous agent needs to partner with a human player, this challenge of effective information exchange becomes more pronounced. Humans typically rely on natural language for communication, a medium that is rich, nuanced, and inherently more complex than the structured data formats that autonomous agents might use. Therefore, successful collaboration between humans and agents often requires more than communication in structured representations---it demands communication in natural language. This requirement introduces an additional layer of complexity, as agents must be able to understand and extract contextually relevant information from natural language communication.

\begin{figure}
    \centering
    \includegraphics[width=\linewidth]{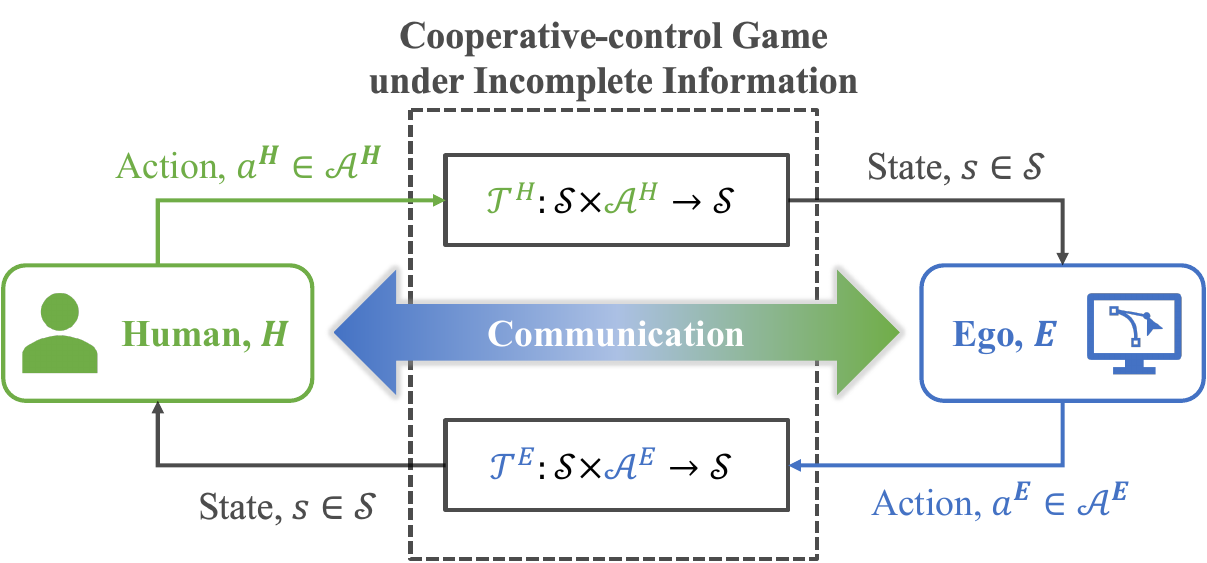}
    \caption{An illustration of the shared-control game in \Cref{def:cooperative_control_game}.}
    \label{fig:shared_control_game}
\end{figure}

Consider a motivating scenario: an autonomous robot and a human coordinator team up in a search-and-rescue mission, navigating hazardous environments to reach a target.
Both the robot and the human have incomplete information: the robot has sensor data of its immediate environment but lacks overall mission context, while the coordinator has an overview plan but no access to the robot's detailed sensor data.

\textit{Gnomes at Night}, a cooperative search-and-find maze board game, offers a more manageable scenario with similar characteristics \cite{gnomesAtNight2016}.
In this board game, two players collaborate to move two magnetically connected gnome pieces through a maze board to collect treasures. The maze paths are different on each side of the board, and players can only move the gnome pieces along paths visible on their side. While they cannot move the gnome pieces through walls on their own side, they can move them through walls on the opposite side. Both players receive identical rewards when reaching a treasure.

To model the described scenarios, we introduce a two-player, turn-based game called a \textit{shared-control game}, where players collectively control a single \textit{token} for a common objective under incomplete information.
This game features a shared state space and individual action spaces. Players take turns to move the token to new states according to their private deterministic transition functions, reflecting their respective understanding of the game dynamics. 
It is worth highlighting that each player does not know the transition function of the other player, creating a form of incomplete information different from what is typically encountered due to partial game state observations as seen in decentralized partially-observable Markov decision processes \cite{bernstein2002complexity}.
Finally, we augment the game with a reward function to incentivize player cooperation in achieving the common objective.
This game abstracts specific scenarios, offering a general framework to think about teamwork in situations under incomplete information.

In this paper, we pose the following problem in shared-control games: 
Knowing the other player is a human, how can an autonomous agent make strategic decisions to achieve smooth cooperation despite lacking complete information on the game dynamics?
We hypothesize that allowing players to exchange information via natural language communication could narrow the information gap between them and lead to more efficient cooperation than without communication.

\begin{figure}[b]
    \centering
    \includegraphics[width=\linewidth]{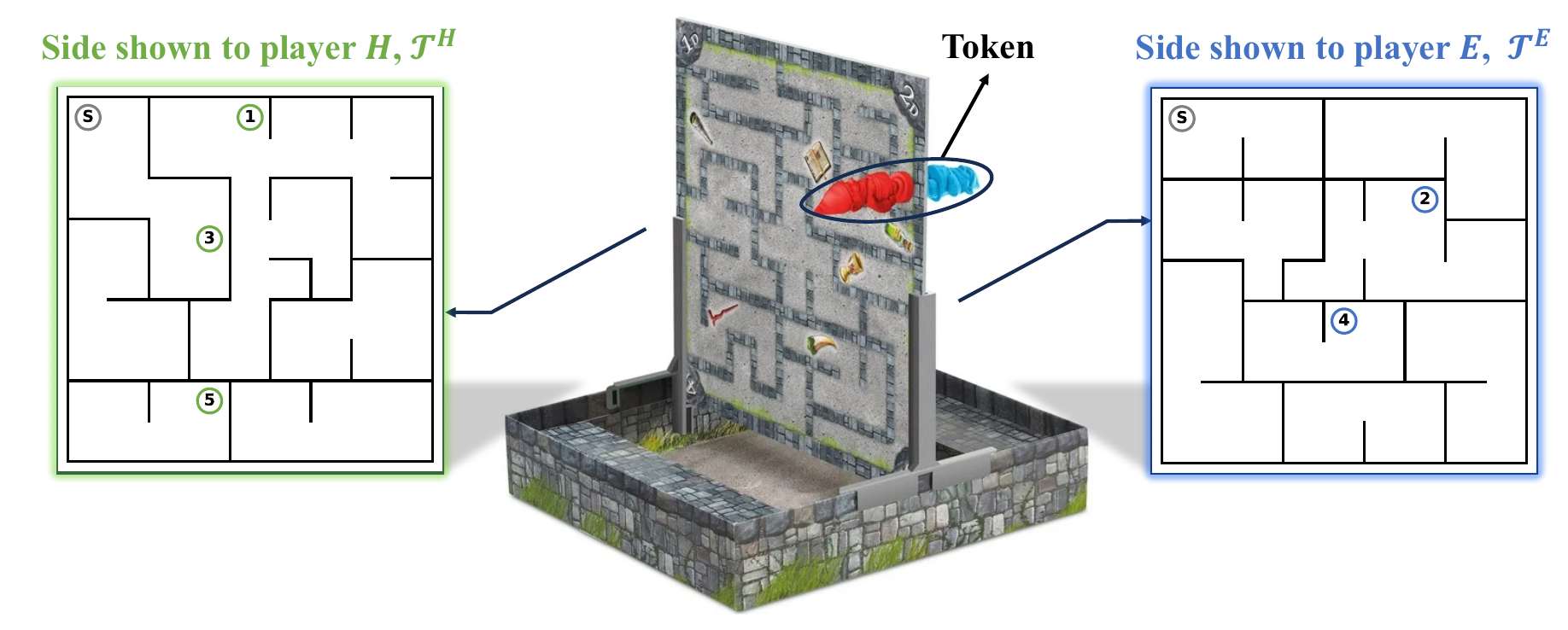}
    \caption{Gnomes at Night testbed where the token is the magnetically connected gnome pieces and two private transition functions encode the wall layouts on each side of the board. Middle figure credit to \protect\cite{gnomesAtNight2016}.}
    \label{fig:gnomes_at_night}
\end{figure}

To test this hypothesis, we introduce a simple testbed based on Gnomes at Night. This testbed is a discretized version of the board game where the maze is a $9\times 9$ gridworld, and players move in single steps. As illustrated in \Cref{fig:gnomes_at_night}, the token is the magnetically connected gnome pieces, and two players see board sides with different wall layouts. This testbed has five rounds, each featuring a different treasure location known exclusively to one player.

We propose an approach that uses natural language communication to inform an online planning algorithm. This approach includes language and planning modules interfaced through a representation of player intents called \textit{flags}.
The language module translates communication messages into flags using a large language model. The planning module then determines the next action based on the input flag and the current game state and optionally generates a new flag.

We present the \textit{asymmetric information-set Monte Carlo tree search with flag exchange} (AISMCTS-F) algorithm, which determines the flag-based policy for the planning module.
This algorithm constructs perspective-based decision trees for each player in a turn-based manner.
An ego player only selects valid actions on its turn, and it assumes initially all actions are valid for the other player since it does not know the other player's transition function.
A \textit{hidden information dictionary}, stored by the ego player, records actions the other player has rejected to improve subsequent decision-making by avoiding these actions.
AISMCTS-F also prioritizes actions that match the other player's communicated preferences, indicated by an input flag, when faced with equally optimal choices.
Lastly, the algorithm expresses its current best estimate of what next action the other player should take via an output flag.

To assess the effectiveness of the proposed approach, we conduct human subject experiments in the Gnomes at Night testbed.
We recruit human participants to play with either another human, a communication-enabled agent implemented via the proposed approach, or an agent without communication capabilities.
Results indicate that enabling communication enhances human-agent cooperation efficiency, reaching treasures in fewer turns and less time per round.
However, even with improvements via the proposed approach, human-agent cooperation still falls short of the efficiency seen in human-to-human gameplay, suggesting potential room for future research. Additionally, We observe distinct communication patterns in human-agent interactions compared to those between humans.



%% file: sections/2_model_and_problem.tex
\section{Cooperation in Shared-control Games under Incomplete Information}\label{sec:game_model}

To model interactions in scenarios described in \Cref{sec:intro}, we introduce a two-player, turn-based game, called a \textit{shared-control game}, where players need to collectively control a single \textit{token} to achieve a common objective under incomplete information. Within the disaster response context, the token is the physical robot, and the two players are the robot's decision-making system and the human operator respectively. 

\begin{definition}\label{def:cooperative_control_game}
A shared-control game where player $\ego$ and player $\human$ collectively controls a token in alternating turns is a tuple $\Gamma = (\Scal, s_{\text{init}}, s_{\text{final}}, \Acal^\ego, \Acal^\human, \Tcal^\ego, \Tcal^\human)$ where $\Scal$ is a finite set of states, $s_{\text{init}}\in\Scal$ is an initial state, $s_{\text{final}}\in\Scal$ is a final state, $\Acal^i$ is a finite set of actions available to each player $i\in\{\ego, \human\}$, and $\Tcal^i:\Scal\times \Acal^i\to \Scal$ is a deterministic transition function for each player $i\in\{\ego, \human\}$.
We refer to player $\ego$ as the \textit{ego player}.
\end{definition}

The shared-control game $\Gamma$ starts with the token in the initial state $s_{\text{init}}\in\Scal$. 
Players $\ego$ and $\human$ take turns to move the token. Either player can initiate the game by taking an action during the first turn.
During the turn of a player $i$, whom we refer to as the \textit{player in control}, it observes the current state of the token in $s\in\Scal$ and selects an action $a$ from its own action space $\Acal^i$.
Then, the token moves to a new state $s'$ according to the transition function $\Tcal^{i}(s,a)$ of this player in control.
Upon transitioning to the new state $s'$, control passes to the other player, i.e., if $\human$ just played, then $\ego$ will take the next turn, and vice versa.
The game continues with players alternating turns until a final state $s_{\text{final}}\in\Scal$ is reached.

Without loss of generality, assume player $\human$ initiates the game, a path through $\Gamma$ is a sequence \(s_0 \raisebox{-1.5pt}{$\xrightarrow{\scriptstyle a^\human_1}$} s_1 \raisebox{-1.5pt}{$\xrightarrow{\scriptstyle a^\ego_2}$} \cdots\)such that $s_t\in\Scal$, $a^i_{t+1}\in\Acal^i$, and $s_{t+1}=\Tcal^i(s_t, a^i_{t+1})$. We assume there exists at least one finite path \(s_0 \raisebox{-1.5pt}{$\xrightarrow{\scriptstyle a^\human_1}$} s_1 \raisebox{-1.5pt}{$\xrightarrow{\scriptstyle a^\ego_2}$} \cdots \raisebox{-1.5pt}{$\xrightarrow{\scriptstyle }$} s_T\) where $s_0=s_{\text{init}}$ and $s_T=s_{\text{final}}$.

\paragraph{Incomplete Information.}
In this game, players have incomplete information about the game dynamics. Each player \(i\) only knows its own transition function \(\Tcal^i\) but not that of the other player. In this paper, we also let the final state $s_{\text{final}}$ be visible to only one player.

\paragraph{Common Reward.}
We augment $\Gamma$ with a common reward function $\Rcal:\Scal\times {\cup}_{i\in \{\ego, \human\}} \Acal^{i} \to\mathbb{R}$ that assigns equal real-valued reward $\Rcal(s,a)$ to both players following the execution of action $a$ by either player in state $s$. This function can capture cooperative objectives, such as reaching a final state in as few turns as possible.

In the \textit{Gnomes at Night} testbed, the state space is the maze grid itself, and both players have the same set of actions: \texttt{noop}, \texttt{right}, \texttt{up}, \texttt{left}, \texttt{down}.
Each player only sees one side of the board, captured by a private transition function.
Players receive equal rewards for finding a treasure and incur penalties for hitting walls or taking excessive steps.
This testbed lays the foundation for exploring the problem of cooperative policy synthesis under incomplete information.

\paragraph{Cooperative Policy Synthesis.}
In a shared-control game, we study the problem of computing cooperative policies for an autonomous agent as the ego player $\ego$, with a human as the other player $\human$.
We allow communication between players via the exchange of messages. Let $M$ be a message space, we model the behavior of player $i$ as a message-based policy $\pi^i:\Scal\times M\to\Acal^i\times M$.

\begin{problem}\label{problem:comm}
In a shared-control game $\Gamma$ with a common reward function $\Rcal$, and given a human player whose behavior is denoted as $\pi^\human$, to compute the ego player's policy $\pi^\ego$ that maximizes the cumulative total reward while adhering to the game transitions:
\begin{subequations}
    \begin{flalign}
        \max_{\pi^\ego} \quad
        & \sum_{t=0}^T \Rcal(s_t, a_t) \label{problem:obj}\\
        \subjectto \quad
        & s_0 = s_{\text{init}}, \label{problem:cons_init_state}\\
        & \text{for all $t\ge 0$,}\nonumber\\
        & \begin{cases}
            a^\human_{t+1}, \color{humangreen}m^\human_{t+1}\color{black} = \pi^\human(s_t, \color{egoblue}m^\ego_t\color{black})  &\text{on $\human$'s turn,}\\
            a^\ego_{t+1}, \color{egoblue}m^\ego_{t+1}\color{black} = 
            \pi^\ego(s_t, \color{humangreen}m^\human_t\color{black}) &\text{on $\ego$'s turn,}
          \end{cases}\label{problem:conts_policy}\\
        & s_{t+1} = \Tcal^i(s_t, a^i_{t+1}), \quad\,\, \text{on player $i$'s turn.}
    \end{flalign}
\end{subequations} 
\end{problem}

Here \(m^i_t\) represents the message sent by player $i$ on turn $t$, and $T$ is the turn at which the final state $s_{\text{final}}$ is reached. 
We highlight the messages sent by the ego player and the human player in blue and green respectively. Each player decides its next move considering the message sent by the other player in the last turn.
This problem generalizes a ``mute version", where messages $m_t^i$ remain unspecified for both players throughout all turns, offering a baseline comparison for understanding the value of communication.

%% file: sections/3_related.tex
\section{Related Work}
Research from various fields is relevant to our problem. This section provides a non-exhaustive overview on two fronts.

\paragraph{Cooperative Games.}
Several strategic games have stood out as testing environments for enhancing player cooperation, including the card game \textit{Hanabi} \cite{bard2020hanabi}, the action video game \textit{Overcooked} \cite{carroll2019utility}, and dialogue games involving negotiation (e.g., \textit{Deal-or-No-Deal} \cite{lewis-etal-2017-deal,he-etal-2018-decoupling} and \textit{Diplomacy} \cite{paquette2019no}) and coordination (e.g., \textit{MutualFriends} \cite{he-etal-2017-learning}).
However, \textit{Gnomes at Night} features a unique set of challenges and game dynamics.  
Unlike Overcooked, which operates under complete information, Gnomes at Night, similar to Hanabi, is characterized by incomplete information. However, the nature of this incompleteness is different: it is due to private transition functions in Gnomes at Night instead of partial observation of game states in Hanabi. Additionally, Gnomes at Night requires natural language communication, significantly different from the hint-based communication in Hanabi and the non-verbal coordination in Overcooked. 
Unlike negotiation games, Gnomes at Night focuses on pure cooperation and collaborative problem-solving, without deceit or manipulation. Finally, other dialogue coordination games usually only require coordination at the end of the dialogue, while our setting requires players to cooperate on each turn by sharing control of a single token.

\paragraph{MCTS-based Planning Techniques.}
Monte Carlo tree search (MCTS) is an algorithm that combines tree-based search with Monte Carlo random sampling to efficiently explore and evaluate vast decision spaces \cite{browne2012survey}. MCTS has proven to be effective in competitive strategic games, such as chess \cite{campbell2002deep}, Shogi \cite{silver2018general}, and Go \cite{silver2016mastering,silver2017mastering}, making it ideal as the foundation of our planning module.
However, standard MCTS works best for games with complete information, but we need to deal with games under incomplete information.
Thus, we adopt the information-set MCTS (ISMCTS), which maintains perspective-based trees for each player, whose nodes correspond to players' information sets and edges correspond to moves from that player's viewpoint \cite{cowling2012information}.
Recent studies have also demonstrated the effectiveness of MCTS-based algorithms in cooperative game contexts, such as in Settlers of Catan \cite{szita2010monte}, The Resistance \cite{cowling2015emergent}, and no-press Diplomacy \cite{jacob2022modeling}. Another work enhanced ISMCTS by introducing re-determinizing methods to prevent hidden information leakage for Hanabi \cite{goodman2019re}. These works highlight MCTS's versatility in not only competitive but also cooperative settings.



%% file: sections/4_method.tex
\section{Method}\label{sec:method}

\begin{figure}
    \centering
    \includegraphics[width=\linewidth]{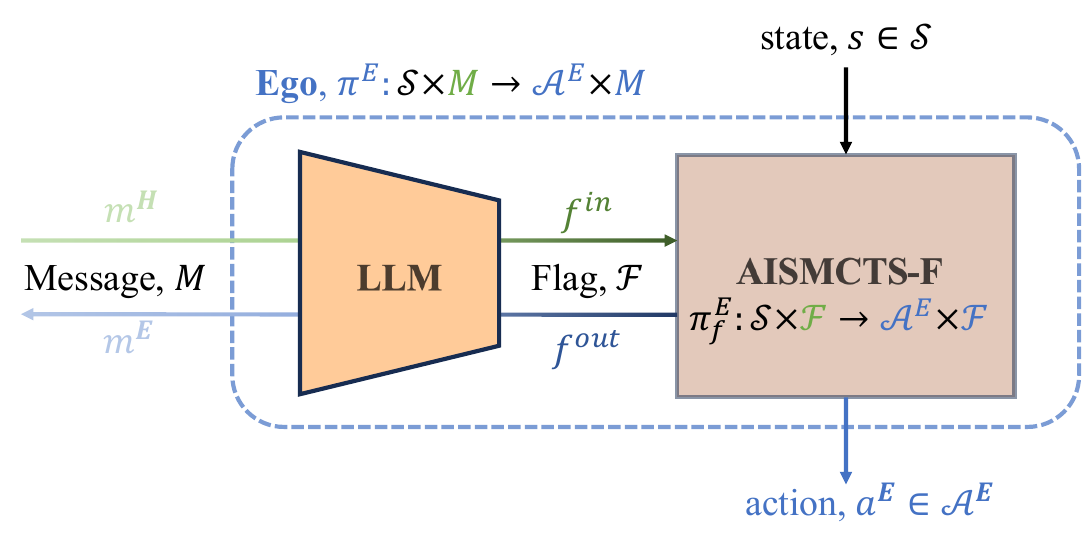}
    \caption{Communication-based approach that generates $\pi^\ego:\Scal\times M\to \Acal^\ego\times M$ via a language module (orange trapezoid) and a planning module (brown rectangle) interfaced through flags.}
    \label{fig:approach_overview}
\end{figure}

This section outlines a communication-based approach that solves \Cref{problem:comm} by leveraging information exchange through natural language communication. It processes a human partner's message and the current state to determine the ego player's next action and optionally generates a responding message. As shown in \Cref{fig:approach_overview}, this approach consists of a language module harnessing a large language model and a planning module based on Monte Carlo tree search, interfaced through a finite set of flags that capture player intents expressed in natural language messages.

We define a \textit{flag} as a compact representation of intents. In a shared-control game, the flag space $\Fcal$ is the union of both players' action spaces and essential responses one player might have to an intent the other player expresses, including acceptance, rejection, inquiry, and no response, i.e., 
\begin{equation}\label{eqn:flag_space}
    \Fcal = \Acal^\human\cup \Acal^\ego\cup\{\texttt{Accept}, \texttt{Reject}, \texttt{Inquiry}, \texttt{None}\}.
\end{equation}
Concretely, the flag space in the Gnomes at Night testbed is
\begin{equation}
\begin{aligned}
    \Fcal = \{&\texttt{noop}, \texttt{right}, \texttt{up}, \texttt{left}, \texttt{down}, \\ 
    &\texttt{Accept}, \texttt{Reject}, \texttt{Inquiry}, \texttt{None}\}.
\end{aligned}
\end{equation}
With this representation defined, we model the ego player's behavior as a flag-based policy \(\pi^\ego_f:\Scal\times \Fcal\to \Acal^\ego \times \Fcal\).

\subsection{Planning Module: AISMCTS-F}
This subsection presents an algorithm for the planning module, called \textit{Asymmetric Information-Set Monte Carlo Tree Search with Flag exchange} (AISMCTS-F), that computes a flag-based policy $\pi^\ego_f$ from the perspective of an ego player $\ego$, detailed in \Cref{algo:AISMCTS-F}.


To explain this algorithm, we introduce these notations: a node $v$ keeps track of its parent $p(v)$, children $C(v)$, incoming action $a(v)$, current state $s(v)$, total reward $T(v)$, and visit count $N(v)$.

Inspired by ISMCTS \cite{cowling2012information}, this algorithm first initializes two trees with root nodes $v_0^i$ for each player $i$. Then the main skeleton of this algorithm (lines \ref{line:main_start}--\ref{line:main_end}) unfolds in four phases:

\begin{algorithm}[t]
    \caption{AISMCTS-F from player $\ego$'s perspective}
    \label{algo:AISMCTS-F}
    \textit{Memory}: hidden information dictionary \(\Omega\), last flag $f^{last}$\\
    \textit{Parameter}: \(\Gamma=(\Scal, s_{\text{init}}, s_{\text{final}}, \Acal^\ego, \Acal^\human, \Tcal^\ego, \Tcal^\human)\), $\Rcal$, $\sigma^\ego$, and the number of iteration \(n\)\\
    \textbf{Input}: current state $s_r\in\Scal$, input flag $f^{in}\in\Fcal$, \\
    \textbf{Output}: action \(a^\star\in\Acal^\ego\), output flag $f^{out}\in\Fcal$
    \begin{algorithmic}[1] 
        \State Create single-node trees with roots $v_0^\ego$ and $v_0^\human$ respectively, and initialize \(\texttt{onRollout} = \text{False}\) \label{line:main_start}
        \For{\(n\) iterations}
            \State Set \(r=0, s=s_r\), denote the \textit{player in control} as $i$
            \While{\(s\ne s_{\text{final}}\)} 
                \State \(a\) = \Call{Explore}{$v^i, s$} \Comment{\textit{Selection}/\textit{Simulation}}
                \State \(r \gets r +\Rcal(s,a)\) 
                \State $s\gets \Tcal^i(s,a)$
                \For{each player $i$} \Comment{\textit{Expansion}}
                    \State $v^i \gets$ \Call{FindOrCreateChild}{$v^i$, $a$}
                \EndFor
                \State \(i\gets \text{the other player}\) 
            \EndWhile
            \For{each player $i$} \Comment{\textit{Backpropagation}}
                \State \Call{Backpropagate}{$v^i$, $r$}
            \EndFor
            \State Reset \(v^\ego\gets v^\ego_0, v^\human\gets v^\human_0\)
            \State Set \(\texttt{onRollout} \gets \text{False}\)
        \EndFor
        \State \textbf{return} \Call{SelectBestAction}{$v^\ego_0, f^{in}$} \label{line:main_end}
    \Statex \rule{\linewidth}{0.5pt} 

    \Function{FindOrCreateChild}{$v$, $a$}
        \If{not \texttt{onRollout}}
            \State Instantiate a node $u$ whose $p(u)=v, a(u)=a$
            \If{$u \notin C(v)$}
                \State Add $u$ to $C(v)$
                \State Set $v\gets u$
                \State Set $\texttt{onRollout} \gets \text{True}$
            \EndIf
            \State \textbf{return} $u$
        \EndIf
    \EndFunction
    \Statex \rule{\linewidth}{0.5pt} 
    
    \Function{Backpropagate}{$v$, $r$}
        \State \(N(v) \gets N(v) +1\)
        \State \(T(v) \gets T(v) + r\)
        \If{$p(v)$ exists} 
            \State \Call{Backpropagate}{$p(v)$, $T(v)$}
        \EndIf
    \EndFunction    
    \end{algorithmic}
\end{algorithm}


\begin{itemize}
    \item \textit{Selection}: For each iteration, the algorithm traverses both trees from the root to a leaf node by selecting nodes that maximize the upper confidence bound ($\ucb$) value \cite{kocsis2006bandit}:
    \begin{equation}
        \ucb(v) = \frac{T(v)}{N(v)} + c \cdot \sqrt{\frac{\log(N(p(v)))}{N(v)}},
    \end{equation}
    where \(c\) balances exploration and exploitation\footnote{We use $\ucb$ for its computational efficiency and set $c=\sqrt{2}$.}.
    \item \textit{Expansion}: Upon reaching a leaf node that has not expanded all possible actions, the algorithm chooses one of these unexplored actions at random and adds a new child node for this action in both trees if such a node does not already exist.
    \item \textit{Simulation}: From new nodes, the algorithm simulates playouts using a rollout policy until the game ends.
    \item \textit{Backpropagation}: Then in both trees, the algorithm backpropagates the results to update the statistics of the nodes visited during this iteration.
\end{itemize}
After $n$ iterations\footnote{We set $n=100$ to meet the real-time requirements in this paper.}, the algorithm returns the action associated with a child node that has been visited the most from among children of the root node $v_0^\ego$ in the ego player's tree.

\begin{figure}[t]
    \centering
    \includegraphics[width=\linewidth]{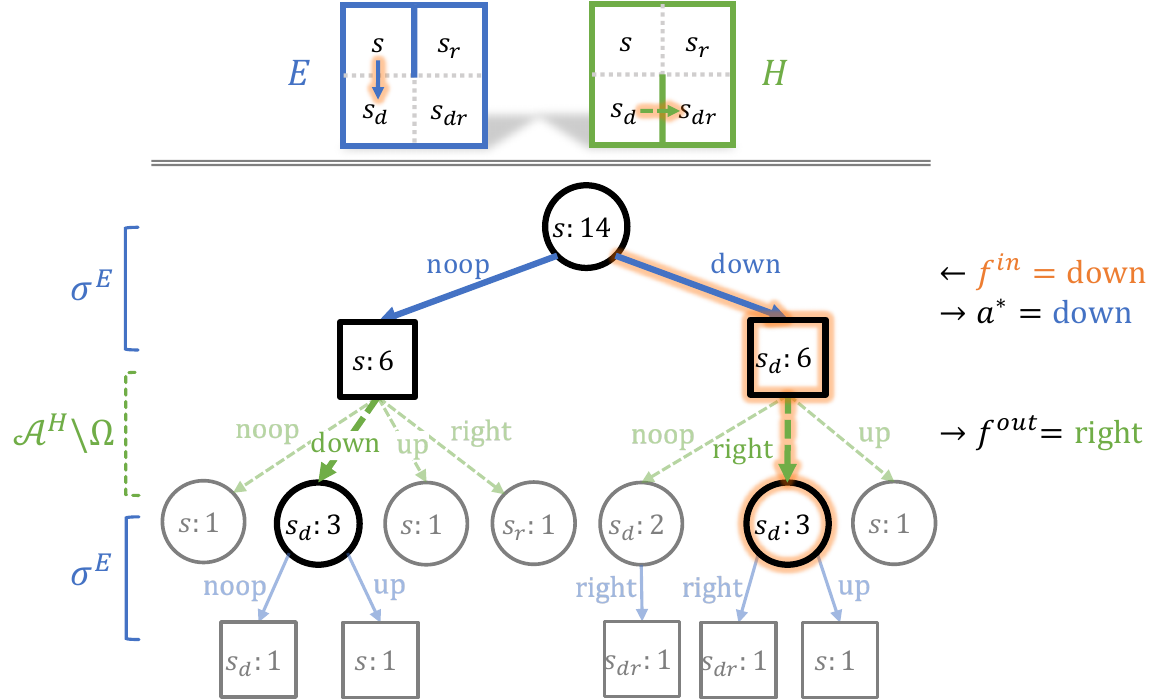}
    \caption{AISMCTS-F: a tree from the perspective of an ego player $\ego$ in a minimal maze example shown at the top.}
    \label{fig:aismctf_ego_tree}
\end{figure}

In the proposed algorithm, we modify two key functions, \textproc{Explore} and \textproc{SelectBestAction}, to utilize the information exchanged in communication. We explain these two functions with a tree from the perspective of an ego player in a minimal maze example, as shown in \Cref{fig:aismctf_ego_tree}.

\paragraph{\textproc{Explore}.}
This function is responsible for choosing an action in both the \textit{Selection} and \textit{Simulation} phases.
In a shared-control game, each player has a private transition function that is inaccessible to the other player.
Each private transition function \(\Tcal^i\) induces a set of valid actions in any given state $s\in\Scal$, which we denote as 
\begin{equation}
    \sigma^{i}(s)=\{a\in \Acal^{i}| \Tcal^{i}(s,a) \text{ is defined}\}.
\end{equation}
A player $\ego$ thus only knows $\sigma^\ego\subseteq \Acal^\ego$ but not $\sigma^\human$, because such sets cannot be defined without the knowledge of $\Tcal^\human$.

This algorithm starts with the ego player assuming all actions available to the other player are valid. Accordingly, as illustrated in \Cref{fig:aismctf_ego_tree}, the ego player's tree expands in an interleaved way: the branching in blue from ego-player-controlled nodes (circles) is less extensive than branching in green from human-player-controlled nodes (squares). This difference occurs because the ego player has less information about the human player's transition function than its own.

We introduce a \textit{hidden information dictionary}, stored in the ego player's memory, as a structure to record the information accumulated through interaction about the transition function of the other player. For player $\ego$, We denote this structure as
\begin{equation}
    \Omega:\Scal\to 2^{\Acal^\human},
\end{equation} 
where $\Omega(s) \subseteq \Acal^\human$ is the set of actions rejected by player $\human$ at state $s$.
In the example shown in \Cref{fig:aismctf_ego_tree}, the algorithm starts the tree-construction with
\begin{equation}
    \Omega = \left\{s:[\texttt{left}], s_d:[\texttt{left}, \texttt{down}]\right\}.
\end{equation}
This piece of information can be useful for reducing the branching factor at the human-player-controlled nodes. Essentially, if the token steps into a visited state where the human has rejected actions $\Omega(s)$, the algorithm will refrain from expanding the corresponding edges as it now believes these actions are unfeasible and not worth exploring. For example, since the human has rejected \texttt{left} at state $s$ and {\texttt{left}, \texttt{down}} at state $s_d$, these corresponding edges are absent in the example tree. In both the \textit{Selection} and \textit{Simulation} phases, we use the hidden information dictionary via \(\Acal^\human\setminus\Omega(s)\) (see \cref{line:instantiate_untried_actions,line:rollout}).

\begin{algorithm}[t]
    \caption*{\textbf{\Cref{algo:AISMCTS-F}} (continued)}
    \begin{algorithmic}[1]
    \makeatletter
    \setcounter{ALG@line}{25} 
    \makeatother    
    \Function{Explore}{$v^i$, $s$}
        \If{not \texttt{onRollout}}\Comment{\textit{Selection}}
            \If{$\Sigma(v^i, s)$ is not instantiated}
                \State \(\Sigma(v^i, s)=\begin{cases}
                \Acal^\human\setminus \Omega(s), \quad &\text{if }i=\human, \\ 
                \sigma^\ego(s), \quad &\text{if }i=\ego.\end{cases}\) \label{line:instantiate_untried_actions}
            \EndIf
            \If{$\Sigma(v^i, s) \ne \emptyset$}
                \State \textbf{return} $a\gets \text{pop}(\Sigma(v^i, s))$
            \EndIf
            \State \textbf{return} \(a(c^*)\) where \(c^*=\arg\max_{c\in C(v^i)} \ucb(c)\) 
        \Else\Comment{\textit{Simulation}}
            \State \textbf{return} \(a\in_{\text{random}}\begin{cases}
            \Acal^\human\setminus\Omega(s),  &\text{if }i=\human, \\ 
            \sigma^\ego(s),  &\text{if }i=\ego.\end{cases}\) \label{line:rollout}
        \EndIf
    \EndFunction
    \Statex \rule{\linewidth}{0.5pt} 
    \Function{SelectBestAction}{$v^\ego_0$, $f^{in}$}
        \If{$f^{in}=\texttt{Inquiry}$} \label{line:phase1_start}
            \State $f^{out}=\texttt{Inquiry}$
        \ElsIf{$f^{in}=\texttt{REJECT}$}
            \State Add $f^{last}$ to $\Omega(s(v^\ego_0))$ 
        \ElsIf{$f^{in}\notin \sigma^\ego(s(v^\ego_0))$}
            \State $f^{out}=\texttt{Reject}$
        \EndIf\label{line:phase1_end}
        \State Compute $\mathcal{C} = \argmax{c\in C(v^\ego_0)}N(c)$ \label{line:most_visit_children_set} \label{line:phase2_start}
        \If{$\exists c^f\in \mathcal{C}$ s.t. $a(c^f)=f^{in}$}
            \State $a^\star \gets f^{in}$, $c^\star \gets c^f$
        \Else
            \State Choose $c^\star\in \mathcal{C}$ , $a^\star\gets a(c^\star)$
        \EndIf\label{line:phase2_end}
        \If{$f^{out}$ not already set} \label{line:phase3_start}
            \State Compute \(\mathcal{G}=
                \{g\in C(c^\star)|a(g)\notin \Omega(s(c^\star))\}\) \label{line:choose_grandchild}
            \If{$\mathcal{G}=\emptyset$}
                \State $f^{out} = \texttt{NONE}$
            \Else
                \State $f^{out} = a(g^\star)$ where $g^\star\in\argmax{g\in \mathcal{G}} N(g)$
            \EndIf\label{line:phase3_end}
        \EndIf
        \State \textbf{return} \(a^\star, f^{out}\), and set $f^{last}\gets f^{out}$
    \EndFunction
    \end{algorithmic}
\end{algorithm}

\paragraph{\textproc{SelectBestAction}.}
This function realizes the flag exchange in the algorithm: given the root node whose state $s(v_0^\ego)$ is the current state in gameplay and an input flag $f^{in}\in\Fcal$, this function operates in three stages to compute the next action $a^\star$ and an output flag $f^{out}\in\Fcal$.

In the initial stage (lines \ref{line:phase1_start}--\ref{line:phase1_end}), the function processes three flag types. For \texttt{Inquiry}, it recognizes a question from the human player, setting the output flag $f^{out}$ to \texttt{Inquiry} too and directing the language module to formulate a response considering the current game information. When encountering \texttt{Reject}, it interprets the human player's response as a refusal of the last proposed action $f^{last}$ and updates the hidden information dictionary accordingly. Lastly, if faced with an invalid action, it issues a \texttt{Reject} flag, prompting the language module to inform the human player of an obstructing wall, conveying hidden information from the ego player to the human player.

The second stage (lines \ref{line:phase2_start}--\ref{line:phase2_end}) in the function selects a node $c^\star$ from among the root node's children and returns the action leading to that node $a^\star$ as the ego player's next move.
The function first finds out a set of child nodes with the highest visit count as $\mathcal{C}$ (see \cref{line:most_visit_children_set}). The actions leading up to these nodes are equally optimal. 
Input flags can serve to break ties, steering the ego player towards an action preferred by the human. For instance, in \Cref{fig:aismctf_ego_tree}, when both \texttt{noop} and \texttt{down} lead to nodes with equal visit counts, an input flag favoring \texttt{down} would guide the selection towards this action, highlighted by a path in orange glow.
Absent a relevant input flag, the function may randomly select a node among $\mathcal{C}$ and return the action leading to this node.

In the final stage (lines \ref{line:phase3_start}--\ref{line:phase3_end}), if the output flag $f^{out}$ is neither set to \texttt{Inquiry} nor \texttt{Reject}, the function generates an output flag based on an action associated with a child node of the previously chosen node $c^\star$, pruning any nodes with actions listed in the hidden information dictionary at $c^\star$'s state (see \cref{line:choose_grandchild}). 
The output flag is the action leading to the highest visit count node in the pruned list, given such a list is not empty.
In the minimal example shown in \Cref{fig:aismctf_ego_tree}, $f^{out}$ is \texttt{right}, which says the ego player wants the other player to move right next. However, as this move is in fact not allowed in its maze, player $\human$ sends a \texttt{Reject} flag. The algorithm will then update the hidden information dictionary to include \texttt{right} at state $s_d$.

In the worst-case scenario, the runtime of \Cref{algo:AISMCTS-F} is $O\left(n(l+\log|\Scal| + |\Acal^\ego| + |\Acal^\human|)\right)$, where $n$ is the number of iterations and $l$ is the average length of simulations.

\subsection{Language Module}
For parsing a human message into an intent flag, we first handle cases of empty or absent messages by returning a \texttt{None} flag. Subsequently, we employ few-shot prompting \cite{brown2020language} with \texttt{gpt-3.5-turbo} to categorize messages into flags. Our prompts use $6$ examples that pair messages and intent flags, with the specific prompt detailed in the supplementary material.

To generate messages based on an intent flag, we use templated sentences as well as GPT-generated responses.
For action flags, our messages request human interaction using the template \textit{``Can you \{\}?"}.
For \texttt{Reject} flags, which indicates an action proposed by the human player is invalid for the ego player at the current state, we notify the human with the template \textit{``I cannot \{\} because there is a wall in that direction."} to explain the obstruction.
For \texttt{Inquiry} flags, we call \texttt{gpt-3.5-turbo} with current game information to generate responses to specific questions. Implementation details for this are also available in the supplementary material.

%% file: sections/5_experiments.tex
\section{Human Subject Experiment}\label{sec:exp}
To assess the effectiveness of the proposed communication-based approach, we conduct human subject experiments with the following hypotheses:
\begin{itemize}
    \item [\textbf{H1.}] Using a large language model for inferring intents is more accurate than for predicting next actions directly.
    \item[\textbf{H2.}] Enabling information exchange via natural language communication improves cooperation efficiency.
\end{itemize}

\begin{figure*}
    \centering
    \begin{subfigure}{0.48\textwidth}
        \centering
        \includegraphics[width=\linewidth]{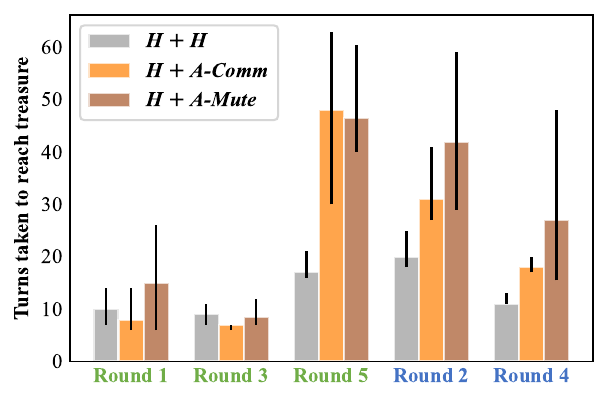}
        \caption{Turns taken.}
        \label{fig:turns}
    \end{subfigure}
    \hfill 
    \begin{subfigure}{0.48\textwidth}
        \centering
        \includegraphics[width=\linewidth]{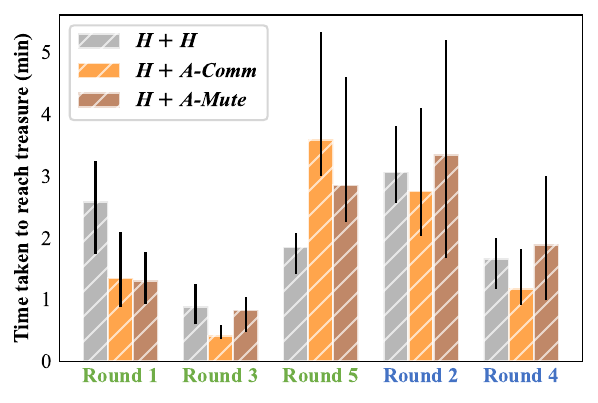}
        \caption{Time taken (min).}
        \label{fig:time}
    \end{subfigure}
    \caption{Bar plots of median cooperation efficiency measures, with 95\% confidence intervals from bootstrap resampling with 10,000 samples.
    Round $1,3,5$ (green): treasure visible to human player; round $2,4$ (blue): treasure visible to ego player (see \Cref{fig:gnomes_at_night}).}
    \label{fig:cooperation-efficiency-measures}
\end{figure*}


\paragraph{Independent Variables.} 
We recruit participants to play as the human player and vary the ego player among three types: another human (\textbf{$\humanIt$}), a communication-enabled agent (\textbf{$\agentComm$}) implemented with the proposed approach, and a mute agent (\textbf{$\agentMute$}) implemented with the same planning module but without communication.

\paragraph{Dependent Measures.}
We use the number of turns as a proxy measure for cooperation efficiency in this turn-based game and measure the time to complete each round. We also analyze the quantity and length of messages both players sent during gameplay.

\paragraph{Experiment Design.}
We employ a between-subject design, where each participant interacts exclusively with one type of ego player. Upon giving consent, participants are asked to play five rounds of Gnomes at Night on our testbed website, each displaying a different treasure location.

\paragraph{Participants.}
We recruit a total of $150$ consenting participants on Prolific \cite{palan2018prolific} to play with the three types of ego players. The average age is $34.71$, with a gender ratio of $0.53$ female.


\subsection{Results and Discussion}

\paragraph{Regarding H1.} From the human-to-human gameplay data collected, we pick out $286$ pairs of conversation snippets and their corresponding subsequent actions. We then annotate snippets with intent flags.
Also using few-shot prompting, we construct prompts with $10\%$ of the tuples and evaluate our chosen LLM, \texttt{gpt-3.5-turbo}, with the remaining $90\%$ of the tuples. The details are in the supplementary material.
Our evaluation demonstrates that predicting flags achieves an accuracy of $74.32\%$, higher than the $53.31\%$ accuracy of directly predicting the next action. 
This result supports hypothesis H1 and justifies our proposed approach.

\paragraph{Regarding H2.}
\Cref{fig:turns} shows that teams of humans and communication-enabled agents ($\humanIt+\agentComm$) outperform the teams of humans and mute agents ($\humanIt+\agentMute$) in $4$ out of $5$ rounds by taking fewer turns to reach the treasure. 
An ANOVA analysis of three groups revealed a significant difference in the number of turns taken ($F(2, 627)=10.53, p< .01$), with post-hoc Tukey HSD testing confirming that $\humanIt+\agentComm$ complete rounds with fewer turns than $\humanIt+\agentMute$ ($p<.05$). 
These results support hypothesis H2. 

\paragraph{Analysis on Completion Time.}
We also measure the time taken to complete each round. Results in \Cref{fig:time} show $\humanIt+\agentComm$ reaches treasures with less or comparable time than $\humanIt+\agentMute$ in $4$ out of $5$ rounds, highlighting the value of communication considering players do not need to spend time sending messages in the latter case.
The same figure also shows $\humanIt+\agentComm$ outpaces $\humanIt+\humanIt$ in $4$ out of $5$ rounds. However, this may not be a reliable indicator that cooperation between humans and communication-enabled agents reaches the efficiency of human-to-human cooperation, since humans may take more time to type messages.

\begin{figure}[b]
    \centering
    \begin{minipage}{.4\linewidth}
        \centering
        \includegraphics[width=\linewidth]{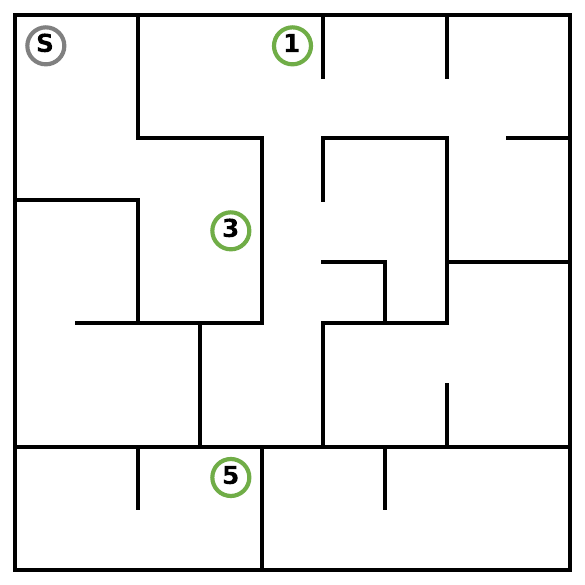}
    \end{minipage}
    \hspace{.05\columnwidth}
    \begin{minipage}{.4\linewidth}
        \centering
        \includegraphics[width=\linewidth]{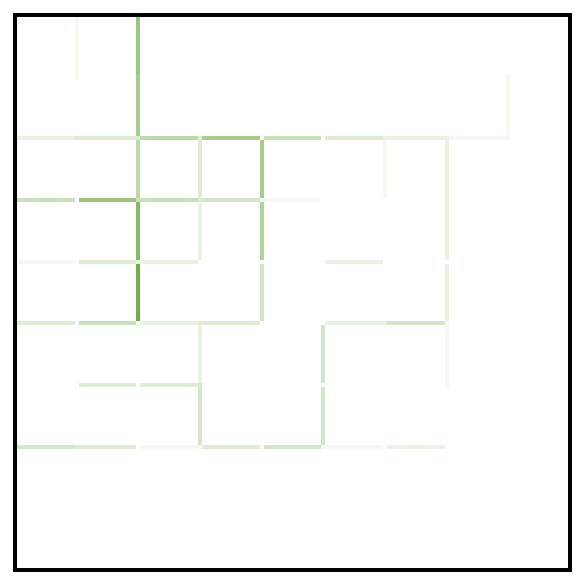}
    \end{minipage}
    \caption{Left: maze side visible to $\humanIt$. Right: heatmap displaying hidden information inferred by $\agentComm$.}
    \label{fig:maze-human-comparison}
\end{figure}

\paragraph{Round $5$ as a Special Case.}
Both turns and time taken results deviate from our hypothesis in round $5$, as depicted in the left maze of \Cref{fig:maze-human-comparison}, where a spanning wall blocks all access to the treasure. 
We speculate that this unique treasure position necessitates more complex strategy coordination over a longer time horizon than what one-step intent communication allows. 
Therefore, human teams significantly outperform both types of human-agent pairs. The slight performance gap between agents that can communicate and those that cannot could simply be due to the time spent on communication.

\paragraph{Communication Paradigms.}
We analyze the average number of messages each player sent during gameplay. The data, presented in \Cref{tab:message}, reveals that humans exchange a similar number of messages with comparable lengths. In contrast, the communication-enabled agent sends around twice as many messages of longer lengths than those sent by its human partner, demonstrating two different communication paradigms. The autonomous agent appears to employ frequent querying as a strategy to obtain more information from its human partner, while human players tend to engage in more equal conversations.

\begin{table}[ht]
    \centering
    \begin{tabular}{lr@{\hspace{1mm}}c@{\hspace{1mm}}rr@{\hspace{1mm}}c@{\hspace{1mm}}r}
        \toprule
        \textbf{Player $\human$ / Player $\ego$}  & \multicolumn{3}{c}{$\humanIt$ / $\humanIt$} & \multicolumn{3}{c}{$\humanIt$ / $\agentComm$} \\
        \midrule
        Average message count     & $7.35$ & / & $7.34$  & $9.24$ & / & $17.57$  \\
        Average message length  & $14.89$ & / & $11.88$  & $11.81$ & / & $25.41$ \\
        \bottomrule
    \end{tabular}
    \caption{Average message count and length per round. $\humanIt + \agentMute$ results are excluded since no messages are allowed.}
    \label{tab:message}
\end{table}

\Cref{fig:maze-human-comparison} qualitatively shows the effectiveness of the proposed approach in learning hidden information via communication.
By visualizing a heatmap of the hidden information dictionary stored at the end of $\humanIt+\agentComm$ gameplays, we show that the ego player successfully discerns the general layout of walls on the human player's side.
Yet, the same figure also reveals instances of mistakenly identified walls, possibly due to human errors or language parsing inaccuracies.

%% file: sections/6_conclusion.tex
\section{Conclusion}
In this paper, we introduce a shared-control game where two players take turns to collectively control a token under incomplete information.
We formulate a policy synthesis problem in this game for an ego player whose partner is a human and hypothesize that communication can narrow the information gap between players for more efficient cooperation.
Our approach combines a language module that translates natural language messages into intent flags and a planning module employing our AISMCTS-F algorithm for informed decision-making.
We conduct human subject experiments where $150$ consenting participants play with either another human, a communication-enabled agent, or a mute agent. 
Results show that the communication-enabled agents tend to outperform mute agents in collaboration with a human player.

\paragraph{Future Work.}
Several experiment results point to directions for future research. First, the deviations in round $5$ suggest future work should extend the communication mechanism to support strategy-making over a longer time horizon. Second, the false positive information in the heatmap highlights the need for methods more robust to human errors and inaccuracies from language parsing. Finally, while this paper adopts a minimal set of intent flags, the proposed approach is a general framework that should work with any type of intent representation, as long as it can be processed by the planning module. Future work should explore representations that can encode richer or composite intents.

%% file: sections/supplementary.tex
\appendix

\section{GPT Prompts}
We use the \texttt{gpt-3.5-turbo} model throughout as our large language model for handling language tasks.

\subsection*{In The Language Module}
The language module leverages the \texttt{gpt-3.5-turbo} model to fulfill two functionalities outlined in Section 4.2 of the paper. 
The first functionality involves parsing human messages into intent flags using few-shot prompting with the following prompt.

\begin{tcolorbox}[
    title={Prompt for parsing {\bf \textcolor{humangreen}{\texttt{message}}}},
    colback=white,
    colframe=gray!70!white,
    colbacktitle=gray!30!white,
    coltitle=black,
    subtitle style={boxrule=0.4pt, colback=yellow!50!white}
    ]
\textcolor{agentmutebrown}{
\text{System Role:} 
\begin{quote}
    Analyze the conversation and predict the next desired move or question. Do not provide explanations or multiple options.
\end{quote}}

\text{Introduction to Task:}
\begin{quote}
    Based on the conversation below, identify the key intention or action. Output MUST be strictly one of the following: [\texttt{noop}, \texttt{right}, \texttt{up}, \texttt{left}, \texttt{down}, \texttt{Accept}, \texttt{Reject}, \texttt{Inquiry}].
\end{quote}

\text{Examples:}
\vspace{\topsep}

\begin{tabular}{@{\hspace{20pt}}lp{9cm}l}
\text{(1)} & \text{Message:} ``Right and then down. First move should be right.'' & \text{Flag:} \texttt{right} \\
\text{(2)} & \text{Message:} ``Can you move left by one step?'' & \text{Flag:} \texttt{left} \\
\text{(3)} & \text{Message:} ``Can you stay put?'' & \text{Flag:} \texttt{noop} \\
\text{(4)} & \text{Message:} ``Ok.'' & \text{Flag:} \texttt{Accept} \\
\text{(5)} & \text{Message:} ``I cannot, there is a wall in that direction.'' & \text{Flag:} \texttt{Reject} \\
\text{(6)} & \text{Message:}``Where exactly is the hidden treasure located?'' & \text{Flag:} \texttt{Inquiry} \\
\end{tabular}
\vspace{\topsep}

\text{New Prompt:}
\renewcommand{\labelitemi}{}
\begin{quote}
    \text{Message:} {\bf \textcolor{humangreen}{\texttt{message}}}\\
    \text{Flag:} 
\end{quote}
\end{tcolorbox}

The second function of the language module involves using \texttt{gpt-3.5-turbo} to generate a suitable response when presented with an \texttt{Inquiry} flag, utilizing current game information as input. The specific prompt employed is detailed below.

\begin{tcolorbox}[
    title={Prompt for generating responses given {\bf \textcolor{humangreen}{\texttt{message}}} and \{\textbf{\texttt{tokenPos}}, \textbf{\texttt{action}}, \textbf{\texttt{treasurePos}}, \textbf{\texttt{treasureSide}}\}},
    colback=white,
    colframe=gray!70!white,
    colbacktitle=gray!30!white,
    coltitle=black,
    subtitle style={boxrule=0.4pt, colback=yellow!50!white}
    ]
\textcolor{agentmutebrown}{
\text{System Role:} 
\begin{quote}
    Assist in a maze game called Gnomes at Night. Use the provided coordinates to focus on the general direction instead of specific next steps. Remind the player of possible obstacles in the way. Limit the response to within 30 words.
\end{quote}}

\text{Task Description:}
\begin{quote}
    You are playing a maze game where the goal is to reach the treasure. The maze's coordinate system starts at [0,0] in the top-left corner, with the x-axis increasing to the right and the y-axis increasing downwards.

    Current Game Information:
    \begin{itemize}
        \item At token position: \{\textbf{\texttt{tokenPos}}\}, I take the move \{\textbf{\texttt{action}}\} now.
        \item \{Treasure position: \textbf{\texttt{treasurePos}}\} if \{\textbf{\texttt{tokenSide}} == 1\} else \{You cannot see the treasure position.\}
    \end{itemize}
\end{quote}

\text{Request:}
\begin{quote}
    Respond to the Inquiry: {\bf \textcolor{humangreen}{\texttt{message}}}
\end{quote}
\end{tcolorbox}

\subsection*{For LLM Evaluation Test}
As discussed in Section 5.1 of the main paper, we pick out $286$ pairs of conversation snippets and their corresponding next actions from the human-to-human gameplay data collected. We then manually annotate each snippet with an intent flag. 

We compare the capability of our chosen LLM, \texttt{gpt-3.5-turbo}, using few-shot prompting to (1) predict intent flags and (2) directly predict the next action. To this end, we designate $10\%$ of the dataset---comprising conversation snippets, their subsequent actions, and annotated intent flags---as a prompt dataset, $\mathcal{D}_{\text{prompt}}$. We then evaluate the model's performance on the remaining $90\%$ of the dataset. For each test, we provide the LLM with a conversation snippet $\tilde{s}$ and compare its predictions against the actual next actions or intent flags. Below is the structure of the prompt we use.

\begin{tcolorbox}[
    title={Prompt for evaluating LLM given the prompt dataset \textcolor{pptblue}{$\mathcal{D}_{\text{prompt}} =\{(s, a, f),\dots\}$} and a test snippet \textcolor{humangreen}{$\tilde{s}$}},
    colback=white,
    colframe=gray!70!white,
    colbacktitle=gray!30!white,
    coltitle=black,
    subtitle style={boxrule=0.4pt, colback=yellow!50!white}
    ]
\textcolor{agentmutebrown}{
\text{System Role:} 
\begin{quote}
    You are analyzing a conversation between two players in a maze game capable of extracting key information and predicting \{the next action/desired next move from partner\}.
\end{quote}}

\text{Task Description:}
\begin{quote}
    Extract key information from the conversation below and predict \{the next move/the intent of the other player\} in the strict format of [\texttt{noop}, \texttt{right}, \texttt{up}, \texttt{left}, \texttt{down}].
\end{quote}

\text{Examples:} For each triple \textcolor{pptblue}{$(s, a, f)$} in \textcolor{pptblue}{$\mathcal{D}_{\text{prompt}}$}:
\begin{quote}
    \text{Snippet:} \textcolor{pptblue}{$s$}\\
    \text{\{Next action/Intent flag\}:}  \{\textcolor{pptblue}{$a$} / \textcolor{pptblue}{$f$}\}
\end{quote}

\text{Evaluate on:}
\begin{quote}
    \text{Snippet:} \textcolor{humangreen}{$\tilde{s}$}\\
    \text{\{Next action/Intent flag\}:}  
\end{quote}
\end{tcolorbox}

Our evaluation demonstrates that predicting intent flags yields an accuracy of $74.32\%$, higher than the $53.31\%$ accuracy of directly predicting the next action. 
This result justifies our approach of employing intent flags as intermediaries between natural language processing and planning modules.

\section{Gnomes at Night Testbed Interface}
We create the Gnomes at Night testbed interface utilizing the multiplayer online game framework \textit{Hathora Builder}, with the backend hosted on \textit{Hathora Cloud} and the frontend developed using a combination of \textit{React} and \textit{Phaser Game}.

\begin{figure}[h]
    \centering
    \begin{minipage}{.45\textwidth}
        \includegraphics[width=\linewidth]{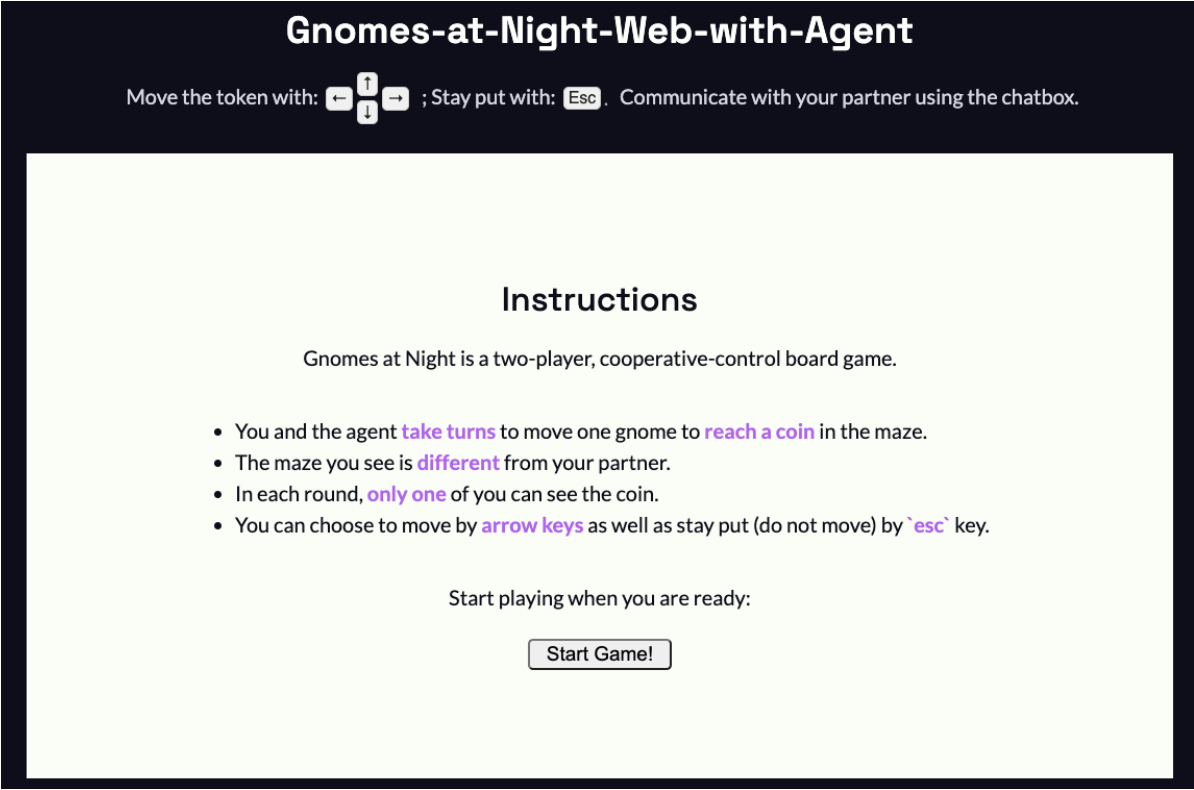}
        \caption{Instructions Page.}
        \label{fig:interface_instruction}
    \end{minipage}
    \hspace{.05\columnwidth}
    \begin{minipage}{.45\textwidth}
        \includegraphics[width=\linewidth]{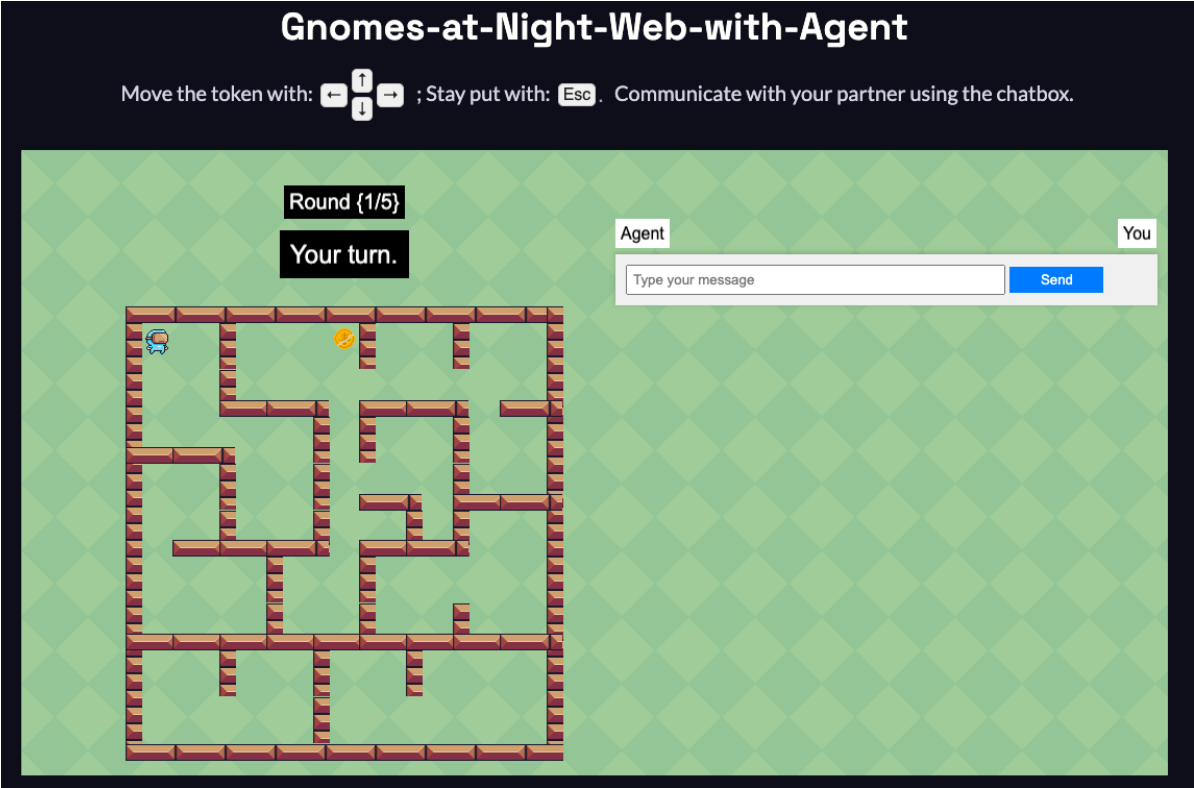}
        \caption{Gameplay Page.}
        \label{fig:interface_gameplay}
    \end{minipage}
\end{figure}

For each type of interaction, participants are initially presented with the game rules on an instruction page (see \Cref{fig:interface_instruction}). After clicking the "Start Game!" button, they are redirected to the gameplay page (\Cref{fig:interface_gameplay}).

\begin{wrapfigure}{r}{0.5\textwidth} 
  \centering
  \includegraphics[width=0.48\textwidth]{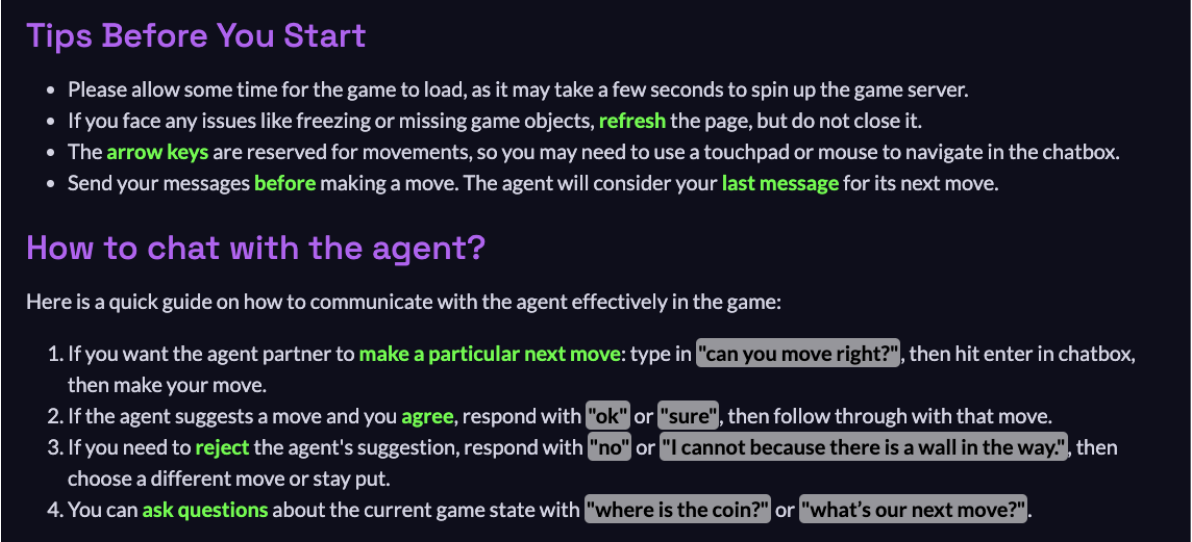}
  \caption{Hints shown in $\humanIt+\agentMute$ interaction.}
  \label{fig:interface_hints}
  \vspace{-3em}
\end{wrapfigure}

The interface varies slightly among the interaction types:
\begin{itemize}
    \item In the $\humanIt+\humanIt$ interaction, the interface does not show the ``Communicate with your partner using the chatbox" prompt in the top bar and excludes the chatbox on the gameplay page.
    \item In the $\humanIt+\agentComm$ interaction, the interface includes a button for participants to copy the site's URL and share it with a friend to join the game.
    \item In the $\humanIt+\agentMute$ interaction, the interface displays an additional list of hints shown in \Cref{fig:interface_hints} that gives participants a better idea of how to communicate with the agent more effectively.
\end{itemize}

\section{Human Subject Experiments}
We have detailed the methodology for the human subject experiment in Section 5 of the paper. Here, we provide a comprehensive breakdown of the participant data we collected for each interaction type and round, along with subjective measures that assess participant engagement and satisfaction.

\subsection*{Participant Data Allocation}
In our human subject experiment, we asked participants to complete all five rounds. However, not all participants followed this instruction, resulting in varying number of gameplay instances for analysis across different rounds and ego player types.

\begin{figure}[h]
    \centering
    \begin{minipage}{.4\textwidth}
        \centering
        \begin{tabular}{
            l
            S[table-format=2.0]
            S[table-format=2.0]
            S[table-format=2.0]
            S[table-format=2.0]
            S[table-format=2.0]
        }
            \toprule
            Round & { $1$} & { $2$} & { $3$} & { $4$} & { $5$} \\
            \midrule
            {$\humanIt+\humanIt$} & {$57$} & {$50$} & {$46$} & {$46$} & {$45$} \\
            {$\humanIt+\agentComm$} & {$45$} & {$42$} & {$40$} & {$40$} & {$39$} \\
            {$\humanIt+\agentMute$} & {$48$} & {$36$} & {$36$} & {$32$} & {$28$} \\
            \bottomrule
        \end{tabular}
        \captionof{table}{Summary of the number of gameplay instances collected for each type of ego player---human ($\humanIt$), agent with communication ($\agentComm$), and agent without communication ($\agentMute$)---over five rounds.}
        \label{tab:num_gameplay}
    \end{minipage}
    \hspace{.01\columnwidth}
    \begin{minipage}{.55\textwidth}
        \centering
        \begin{tikzpicture}
            \begin{axis}[
                width=7cm,
                height=4.5cm,
                xlabel={Round},
                ylabel={Adherence Rate (\%)},
                xlabel style={yshift=0.2em}, 
                ylabel style={yshift=-0.2em}, 
                xmin=0.75, xmax=5.25, 
                ymin=55, ymax=105,
                xticklabels={ $1$,  $2$,  $3$,  $4$,  $5$},
                xtick={1, 2, 3, 4, 5},
                xtick style={draw=none}, 
                ytick={50,60,70,80,90,100},
                ytick style={draw=none}, 
                legend style={at={(1.35,0.82)},anchor=north,legend cell align=left}, 
                ymajorgrids=true,
                grid style=dashed,
                legend entries={$\humanIt+\humanIt$ ($57$),$\humanIt+\agentComm$ ($45$), $\humanIt+\agentMute$ ($48$)},
                ]
            
            \addplot[
                color=humanhumangray,
                line width=1.5pt, 
                mark=square,      
                ]
                coordinates {
                (1,100)(2,87.7)(3,80.7)(4,80.7)(5,78.9)
                };
            
            \addplot[
                color=agentagentorange,
                line width=1.5pt, 
                mark=*,
                ]
                coordinates {
                (1,100)(2,93.3)(3,88.9)(4,88.9)(5,86.7)
                };
            
            \addplot[
                color=agentmutebrown,
                line width=1.5pt, 
                mark=triangle,
                ]
                coordinates {
                (1,100)(2,75)(3,75)(4,66.7)(5,58.3)
                };
            
            \end{axis}
            \label{plot:adherence_rate}
        \end{tikzpicture}
    \end{minipage}
\end{figure}

\Cref{tab:num_gameplay} details the number of gameplay data collected by round and ego player type, showing a drop in participation as the game progresses. 
We calculate the adherence rate for each interaction type in each round by calculating the ratio of the number of participants who completed a given round to that in the initial round. The adherence rate plot on the right demonstrates that fewer than $60\%$ of participants in $\humanIt+\agentMute$ interaction stick to the final round, in contrast to approximately $80\%$ or higher in both $\humanIt+\humanIt$ and $\humanIt+\agentComm$ interactions. 
This suggests that the mute version of the Gnomes at Night treasure-finding problem is very difficult, leading to participant impatience and a higher likelihood of quitting the game before completing all rounds.

\subsection*{Subjective Measures on Engagement and Satisfaction}
Following the completion of the Gnomes at Night testbed gameplay, participants were asked to rate their engagement and satisfaction levels in two 5-point Likert scale questions.
Here, we present these subjective metrics to complement the main metrics of turns taken and time taken highlighted in the paper, to guide future improvements.

\begin{wraptable}{r}{7cm}
    \centering
    \begin{tabular}{lr@{\hspace{1mm}}c@{\hspace{1mm}}rr@{\hspace{1mm}}c@{\hspace{1mm}}r}
        \toprule
         & \multicolumn{3}{c}{Engagement} & \multicolumn{3}{c}{Satisfaction} \\
        \midrule
        $\humanIt+\humanIt$     & $4.78$ & $\pm$ & $0.53$  & $4.45$ & $\pm$ & $1.01$  \\
        $\humanIt+\agentComm$  & $4.47$ & $\pm$ & $0.88$  & $2.83$ & $\pm$ & $1.20$ \\
        $\humanIt+\agentMute$ & $4.56$ & $\pm$ & $0.70$  & $3.22$ & $\pm$ & $1.22$ \\
        \bottomrule
    \end{tabular}
    \captionof{table}{Summary of participant engagement and satisfaction on 5-point Likert scales, presented as mean $\pm$ standard deviation.}
    \label{tab:subject_measures}
\end{wraptable}

\Cref{tab:subject_measures} shows a summary of average engagement and satisfaction scores among participants across three interaction types. 
As anticipated, interactions between humans ($\humanIt+\humanIt$) scored the highest in both measures.
Surprisingly, average engagement and satisfaction scores for $\humanIt+\agentComm$ are lower than those for $\humanIt+\agentMute$, which is puzzling as we would assume giving the participants the channel of communication increases such measures. The result, however, shows the reverse effect. 
One intuition is that participants may have higher expectations for agent capabilities when communication is possible, suggesting room for improvement in the language module to meet these expectations.

To further examine these results, we performed an ANOVA analysis on the engagement and satisfaction data to determine if the differences in mean values are statistically significant.
For engagement, the ANOVA analysis does not reveal a significant difference among the three interaction types ($F(2,104)=1.840, p=0.164$).
For satisfaction, the ANOVA analysis reveals a significant difference among three interaction types ($F(2, 104)=21.790, p<.01$). However, the post-hoc Tukey HSD test identified differences only between $\humanIt+\humanIt$ and both $\humanIt+\agentComm$ ($p<.01$) and $\humanIt+\agentMute$ ($p<.01$), which is as expected.
These statistical tests indicate that the observed differences might not be as robust as presumed, which is also why we chose not to highlight these findings in the main paper.




%% file: ijcai24.bbl
\begin{thebibliography}{}

\bibitem[\protect\citeauthoryear{Bard \bgroup \em et al.\egroup }{2020}]{bard2020hanabi}
Nolan Bard, Jakob~N Foerster, Sarath Chandar, Neil Burch, Marc Lanctot, H~Francis Song, Emilio Parisotto, Vincent Dumoulin, Subhodeep Moitra, Edward Hughes, et~al.
\newblock {The Hanabi challenge: A new frontier for AI research}.
\newblock {\em Artificial Intelligence}, 280:103216, 2020.

\bibitem[\protect\citeauthoryear{Bernstein \bgroup \em et al.\egroup }{2002}]{bernstein2002complexity}
Daniel~S Bernstein, Robert Givan, Neil Immerman, and Shlomo Zilberstein.
\newblock {The complexity of decentralized control of Markov decision processes}.
\newblock {\em Mathematics of Operations Research}, 27(4):819--840, 2002.

\bibitem[\protect\citeauthoryear{Brown \bgroup \em et al.\egroup }{2020}]{brown2020language}
Tom Brown, Benjamin Mann, Nick Ryder, Melanie Subbiah, Jared~D Kaplan, Prafulla Dhariwal, Arvind Neelakantan, Pranav Shyam, Girish Sastry, Amanda Askell, et~al.
\newblock {Language models are few-shot learners}.
\newblock In {\em Advances in Neural Information Processing Systems}, volume~33, pages 1877--1901, 2020.

\bibitem[\protect\citeauthoryear{Browne \bgroup \em et al.\egroup }{2012}]{browne2012survey}
Cameron~B Browne, Edward Powley, Daniel Whitehouse, Simon~M Lucas, Peter~I Cowling, Philipp Rohlfshagen, Stephen Tavener, Diego Perez, Spyridon Samothrakis, and Simon Colton.
\newblock {A survey of Monte Carlo tree search methods}.
\newblock {\em Transactions on Computational Intelligence and AI in Games}, 4(1):1--43, 2012.

\bibitem[\protect\citeauthoryear{Campbell \bgroup \em et al.\egroup }{2002}]{campbell2002deep}
Murray Campbell, A~Joseph Hoane~Jr, and Feng-hsiung Hsu.
\newblock {Deep Blue}.
\newblock {\em Artificial intelligence}, 2002.

\bibitem[\protect\citeauthoryear{Carroll \bgroup \em et al.\egroup }{2019}]{carroll2019utility}
Micah Carroll, Rohin Shah, Mark~K Ho, Tom Griffiths, Sanjit Seshia, Pieter Abbeel, and Anca Dragan.
\newblock {On the utility of learning about humans for human-AI coordination}.
\newblock In {\em Advances in Neural Information Processing Systems}, 2019.

\bibitem[\protect\citeauthoryear{Cowling \bgroup \em et al.\egroup }{2012}]{cowling2012information}
Peter~I Cowling, Edward~J Powley, and Daniel Whitehouse.
\newblock {Information set Monte Carlo tree search}.
\newblock {\em Transactions on Computational Intelligence and AI in Games}, 4(2):120--143, 2012.

\bibitem[\protect\citeauthoryear{Cowling \bgroup \em et al.\egroup }{2015}]{cowling2015emergent}
Peter~I Cowling, Daniel Whitehouse, and Edward~J Powley.
\newblock {Emergent bluffing and inference with Monte Carlo tree search}.
\newblock In {\em Conference on Computational Intelligence and Games}, pages 114--121, 2015.

\bibitem[\protect\citeauthoryear{Goil \bgroup \em et al.\egroup }{2013}]{goil2013using}
Aditya Goil, Matthew Derry, and Brenna~D Argall.
\newblock {Using machine learning to blend human and robot controls for assisted wheelchair navigation}.
\newblock In {\em International Conference on Rehabilitation Robotics}, pages 1--6, 2013.

\bibitem[\protect\citeauthoryear{Goodman}{2019}]{goodman2019re}
James Goodman.
\newblock {Re-determinizing information set Monte Carlo tree search in Hanabi}.
\newblock {\em ArXiv preprint arXiv:1902.06075}, 2019.

\bibitem[\protect\citeauthoryear{He \bgroup \em et al.\egroup }{2017}]{he-etal-2017-learning}
He~He, Anusha Balakrishnan, Mihail Eric, and Percy Liang.
\newblock Learning symmetric collaborative dialogue agents with dynamic knowledge graph embeddings.
\newblock In {\em Annual Meeting of the Association for Computational Linguistics (Volume 1: Long Papers)}, 2017.

\bibitem[\protect\citeauthoryear{He \bgroup \em et al.\egroup }{2018}]{he-etal-2018-decoupling}
He~He, Derek Chen, Anusha Balakrishnan, and Percy Liang.
\newblock {Decoupling strategy and generation in negotiation dialogues}.
\newblock In {\em Conference on Empirical Methods in Natural Language Processing}, 2018.

\bibitem[\protect\citeauthoryear{Jacob \bgroup \em et al.\egroup }{2022}]{jacob2022modeling}
Athul~Paul Jacob, David~J Wu, Gabriele Farina, Adam Lerer, Hengyuan Hu, Anton Bakhtin, Jacob Andreas, and Noam Brown.
\newblock {Modeling strong and human-like gameplay with KL-regularized search}.
\newblock In {\em International Conference on Machine Learning}, pages 9695--9728, 2022.

\bibitem[\protect\citeauthoryear{Kocsis and Szepesv{\'a}ri}{2006}]{kocsis2006bandit}
Levente Kocsis and Csaba Szepesv{\'a}ri.
\newblock {Bandit based Monte-Carlo planning}.
\newblock In {\em European Conference on Machine Learning}, pages 282--293, 2006.

\bibitem[\protect\citeauthoryear{Lewis \bgroup \em et al.\egroup }{2017}]{lewis-etal-2017-deal}
Mike Lewis, Denis Yarats, Yann Dauphin, Devi Parikh, and Dhruv Batra.
\newblock {Deal or no deal? End-to-end learning of negotiation dialogues}.
\newblock In {\em Conference on Empirical Methods in Natural Language Processing}, 2017.

\bibitem[\protect\citeauthoryear{Palan and Schitter}{2018}]{palan2018prolific}
Stefan Palan and Christian Schitter.
\newblock {Prolific. ac—A subject pool for online experiments}.
\newblock {\em Journal of Behavioral and Experimental Finance}, 17:22--27, 2018.

\bibitem[\protect\citeauthoryear{Paquette \bgroup \em et al.\egroup }{2019}]{paquette2019no}
Philip Paquette, Yuchen Lu, Seton~Steven Bocco, Max Smith, Satya O-G, Jonathan~K Kummerfeld, Joelle Pineau, Satinder Singh, and Aaron~C Courville.
\newblock {No-press diplomacy: Modeling multi-agent gameplay}.
\newblock In {\em Advances in Neural Information Processing Systems}, 2019.

\bibitem[\protect\citeauthoryear{{Peaceable Kingdom}}{2016}]{gnomesAtNight2016}
{Peaceable Kingdom}.
\newblock Gnomes at night.
\newblock \url{https://www.walmart.com/ip/Peaceable-Kingdom-Gnomes-at-Night-Game-2-to-4-Players-Ages-6/585042377}, 2016.
\newblock Accessed: 2024-01-15.

\bibitem[\protect\citeauthoryear{Silver \bgroup \em et al.\egroup }{2016}]{silver2016mastering}
David Silver, Aja Huang, Chris~J Maddison, Arthur Guez, Laurent Sifre, George Van Den~Driessche, Julian Schrittwieser, Ioannis Antonoglou, Veda Panneershelvam, Marc Lanctot, et~al.
\newblock {Mastering the game of Go with deep neural networks and tree search}.
\newblock {\em Nature}, 529(7587):484--489, 2016.

\bibitem[\protect\citeauthoryear{Silver \bgroup \em et al.\egroup }{2017}]{silver2017mastering}
David Silver, Julian Schrittwieser, Karen Simonyan, Ioannis Antonoglou, Aja Huang, Arthur Guez, Thomas Hubert, Lucas Baker, Matthew Lai, Adrian Bolton, et~al.
\newblock {Mastering the game of Go without human knowledge}.
\newblock {\em Nature}, 550(7676):354--359, 2017.

\bibitem[\protect\citeauthoryear{Silver \bgroup \em et al.\egroup }{2018}]{silver2018general}
David Silver, Thomas Hubert, Julian Schrittwieser, Ioannis Antonoglou, Matthew Lai, Arthur Guez, Marc Lanctot, Laurent Sifre, Dharshan Kumaran, Thore Graepel, et~al.
\newblock {A general reinforcement learning algorithm that masters chess, shogi, and Go through self-play}.
\newblock {\em Science}, 362(6419):1140--1144, 2018.

\bibitem[\protect\citeauthoryear{Szita \bgroup \em et al.\egroup }{2010}]{szita2010monte}
Istv{\'a}n Szita, Guillaume Chaslot, and Pieter Spronck.
\newblock {Monte-Carlo tree search in Settlers of Catan}.
\newblock In {\em Advances in Computer Games}, pages 21--32, 2010.

\end{thebibliography}
